\documentclass[letterpaper, 10 pt, conference]{ieeeconf}  

\IEEEoverridecommandlockouts                              
\usepackage{xcolor}
\usepackage{balance}
\usepackage{graphics} 
\usepackage{epsfig} 
\usepackage{amsmath} 
\usepackage{amssymb}  
\usepackage{listings} 
\usepackage{cite} 

\usepackage[caption=false,font=footnotesize]{subfig}

\newtheorem{remark}{Remark}

\usepackage{algorithm}
\usepackage{algorithmic}

\usepackage[font=small,skip=1pt]{caption}

\usepackage[utf8]{inputenc}

\usepackage{textcomp}
\usepackage{gensymb}

\let\labelindent\relax
\usepackage[shortlabels]{enumitem}

\title{\LARGE \bf
Adaptive Trajectory Planning and Optimization at Limits of Handling
}

\author{Lars Svensson$^{1}$, Monimoy Bujarbaruah$^{2}$, Nitin R. Kapania$^{3}$ and Martin T\"orngren$^{1}$
\thanks{$^{1}$ L.Svensson and M.T\"orngren are with the Department of Mechatronics and Embedded Control Systems, KTH Royal Institute of
Technology, Stockholm, Sweden. {\tt\small \{larsvens,martint\}@kth.se}}%
\thanks{$^{2}$M. Bujarbaruah is with the Model Predictive Control Laboratory at University of California Berkeley, USA. {\tt\small monimoyb@berkeley.edu}}%
\thanks{$^{3}$N. Kapania is with the Dynamic Design Laboratory at Stanford University, USA. {\tt\small nkapania@stanford.edu}}%
}

\begin{document}

\maketitle
\thispagestyle{empty}
\pagestyle{empty}

\begin{abstract}
In this paper, we tackle the problem of trajectory planning and control of a vehicle under locally varying traction limitations, in the presence of suddenly appearing obstacles. 
We employ concepts from adaptive model predictive control for run-time adaptation of tire force constraints that are imposed by local traction conditions. To solve the resulting optimization problem for real-time control synthesis with such time varying constraints, we propose a novel numerical scheme based on Real Time Iteration Sequential Quadratic Programming (RTI-SQP),
which we call Sampling Augmented Adaptive RTI (SAA-RTI). Sampling augmentation of conventional RTI-SQP provides additional feasible candidate trajectories for warm-starting the optimization procedure. Thus, the proposed SAA-RTI algorithm enables real time constraint adaptation and reduces sensitivity to local minima.   
Through extensive numerical simulations we demonstrate that our method increases the vehicle's capacity to avoid accidents in scenarios with unanticipated obstacles and locally varying traction, compared to equivalent non-adaptive control schemes and traditional planning and tracking approaches.

\end{abstract}

\section{Introduction}
\label{sec:introduction}
Automated driving and advanced driver assistance systems technology is developed and deployed around the world as a means of improving safety and mobility. With deployment increasing, the rate at which these systems are exposed to critical traffic situations also increase. Such situations, e.g. a late detected pedestrian in the vehicle path, or an unanticipated lane change by a nearby vehicle require operation at the handling limits of a vehicle to maximize the capacity to avoid potential accidents. Conservative assumptions about the physical capacity of the vehicle reduces the set of considered maneuvers, which may lead to reduced safety of passengers and road users. Also, the physical limitations of the vehicle typically vary in time due to local road and weather conditions, making motion planning and control in critical situations a challenging task.  

Research in motion planning and control of automated road vehicles has matured rapidly in recent years and numerous academic works have presented algorithms for motion planning and control in general driving scenarios \cite{paden2016survey,gonzalez2016review,ziegler2014trajectory,werling2012_IJRRl}. 
Due to computational limitations, the planning and control problem is generally divided into hierarchical levels with gradually decreasing planning horizon and increasing model fidelity. A dynamic model including tire force modelling is typically only used for trajectory tracking \cite{paden2016survey,gonzalez2016review}, whereas local trajectory planning typically uses less sophisticated models, such as the kinematic bicycle model or the point mass model \cite{ziegler2014trajectory,werling2012_IJRRl}. Not being able to precisely represent the dynamic limitations in the local planner presents a potential problem when motion planning close to the dynamic limits of the vehicle in one of two ways. First, over-estimating the dynamic capabilities may lead to poor tracking performance in the controller, possibly resulting in collision. Second, under-estimating the dynamic capability may cause failure to select an available collision free maneuver. Several works have been proposed to mitigate this discrepancy by employing pre-computed motion primitives in the local planner for which the dynamic limitations are considered \cite{svensson2018safe, liniger2015optimization}. However, the dynamic capabilities of road vehicles are typically prone to substantial local variations in terms of the tire-road friction coefficient \cite{rajamani2012algorithms}, rendering pre-computed motion primitives suboptimal or infeasible in most cases. Hence, accurate estimation of dynamic capabilities coupled with adaptive optimal motion planning and control is required to fully utilize the physical capabilities of the vehicle. We hypothesize that adapting to local road conditions and acting optimally with respect to the associated physical limitations will improve the capacity of the vehicle to handle unforeseen critical traffic situations. 

In this paper, we propose an integrated framework for local planning and control of a vehicle, in which the dynamic constraints of the vehicle can be adapted at run-time. We assume a state-of-the-art solution of the tire-road friction estimation   \cite{rajamani2012algorithms, gustafsson1997slip, rajamani2011vehicle}, and focus on subsequent motion planning and control problem only, using the up-to-date friction estimate.  Our method utilizes a combination of state space sampling \cite{howard2008state} and adaptive model predictive control (MPC) \cite{ostafew2014learning, bujarbaruahAvec}, employed in a Real Time Iteration Sequential Quadratic Programming (RTI-SQP) fashion \cite{diehl2005real}.  
\textcolor{black}{The resulting unified trajectory planning and optimization algorithm, which we call Sampling Augmented Adaptive RTI (SAA-RTI), bears the following contributions: 
\begin{enumerate}
    \item A sampling based strategy for traction adaptive motion planning, which incorporates the knowledge of vehicle model and operating constraints. For this planner, the time-varying tire-road friction limitation is handled as a time-varying adaptive input constraint. \label{item:contrib1}
    \item A trajectory optimization scheme based on RTI-SQP for optimizing planned trajectories from \ref{item:contrib1}), in environments with obstacles. The proposed scheme avoids potential infeasibility and local minima, while utilizing full dynamic capabilities of the vehicle via the adaptive input constraints from (\ref{item:contrib1}).
\end{enumerate}
}
We demonstrate through numerical simulations that our SAA-RTI algorithm increases the vehicle's capacity to avoid obstacles in critical situations compared to an equivalent non-adaptive method, as well as a traditional modular planning and tracking scheme.


\section{Related Work}
\label{sec:relatedwork}

Research in motion planning and control at the handling limits is influenced by research in the racing community. Through the use of nonlinear programming, Perantoni et al. \cite{perantoni2014optimal} computes the time-optimal speed profile and racing line for an entire race track, although computational limitations require the trajectories to be computed offline. Kapania and Gerdes \cite{kapania2016sequential} presents an experimentally validated algorithm that reduces computational expense by breaking down the combined lateral/longitudinal vehicle control problem into two sequential subproblems that are solved iteratively.

Another optimization based approach to the autonomous racing problem is to repeatedly solve a Constrained Finite Time Optimal Control (CFTOC) problem  online. Liniger et al. \cite{liniger2015optimization} utilizes the Real Time Iteration Sequential Quadratic Programming (RTI-SQP) paradigm \cite{gros2016linear} to jointly solve the trajectory planning and control problems. Rosolia \cite{rosoliaracing} applied learning MPC to minimize lap completion time, given data from previous laps. Building on the experiences from the racing application, Gray et al. \cite{gray2012predictive} considered motion planning at the handling limits for obstacle avoidance, generating a high-level motion plan from a four-wheel dynamic model and a low-level plan using MPC. Zhang et al. \cite{zhang2018autonomous} re-formulate the collision avoidance constraints in the dual variable space, which results in a smooth (but still, non-convex) optimization problem. A predictive control approach was also utilized by Funke et al. \cite{funke2017collision} and Brown et al. \cite{brown2017safe} to provide collision-free trajectories while maintaining vehicle stability.
A practical drawback of purely optimization-based motion planning techniques pointed out by Ziegler et al. \cite{ziegler2014trajectory} is that they struggle in situations where the motion planning problem contains discrete decision making (e.g. to go left or right of an obstacle). In specific cases \cite{liniger2015optimization, brown2017safe}, this can be remedied by a high level path planner based on a method such as dynamic programming. However, to the best of our knowledge, a generalized solution for this problem without loss of optimality w.r.t. the dynamic capabilities of the vehicle has yet to be presented. 

Hence for practicality, state space sampling methods such as those presented by Howard et al. \cite{howard2008state} are widely used in industry for collision avoidance. The core concept of the method is as follows. A grid is defined in the terminal state of the planning horizon and a set of two point boundary value problems are solved between the initial state and each sampled terminal state, generating a trajectory set. Dynamic constraints are not considered in the generation of the trajectory set. Instead, a dynamic feasibility check is done in conjunction with the collision check for each trajectory. Werling et al extended the method 
by generating the trajectory set in a road-aligned coordinate frame and by introducing a terminal manifold to improve the selection of terminal states \cite{werling2012_IJRRl}. It has been shown that the method is well suited for planning in scenarios including discrete decisions. However, even though it reliably produces feasible maneuvers, they are suboptimal w.r.t. the physical capabilities of the vehicle. 

An intuitive way to reduce suboptimality of the trajectories in the set is to solve the two-point boundary value problem offline, using a dynamic model. This method has been demonstrated successfully in several previous works \cite{liniger2015optimization,svensson2018safe}. However, this approach prohibits online model adaptation, since the trajectories in the pre-computed library are computed based on a static vehicle representation. 

On the other hand, to account for local variations in physical capabilities of the vehicle, we draw from developments in the field of adaptive control. Predictive control under model uncertainty has been well-studied recently \cite{tanaskovic2014adaptive, hewing2018cautious, bujarbaruahAdapFIR, koller2018learning, bujarbaruahCDC}. Such frameworks allow the system to dynamically re-plan safer and more cost efficient trajectories with time, as additional model information available from data is provided to the MPC optimization problem.  We leverage this notion of \emph{adaptive MPC} in our work as well, by utilizing updated vehicle model information to adapt the constraints to account for time-varying traction limitations. With extensive numerical examples, we highlight that this method of recursive model adaptation in MPC improves the capacity to avoid obstacles under time-varying road conditions.

\section{Problem Formulation}
\label{sec:problemformulation}

We tackle the problem of real-time trajectory planning and control of a vehicle at its limits of handling, under time varying \textcolor{black}{traction limitations}. The controller synthesis is done by solving an optimization problem \textcolor{black}{with time varying constraints} in a receding horizon fashion in real time, i.e., a solution is obtained fast enough to accommodate a sufficiently fast replanning rate \cite{paden2016survey}. In the following section we introduce the model and constraints of the optimization problem. 

\subsubsection*{Vehicle Model}
Throughout this paper we consider a dynamic bicycle model expressed in a road aligned coordinate frame. The state propagation is described in \eqref{eq:dyn_model}: 

\begin{subequations} 
\label{eq:dyn_model}
\begin{align}
& \dot{s} = \frac{v_x \cos{(\Delta \psi)} - v_y \sin{(\Delta \psi)}}{1-d \kappa_c}, \\
& \dot{d} = v_x\sin{(\Delta \psi)} + v_y \cos{( \Delta \psi)}, \\
& \Delta \dot{\psi} = \dot{\psi} - \kappa_c \frac{v_x \cos{(\Delta \psi)} - v_y \sin{(\Delta \psi)}}{1-d \kappa_c}, \\
& \ddot{\psi} = \frac{1}{I_z} \left( l_{\textnormal{f}} F_{{y\textnormal{f}}} - l_{\textnormal{r}} F_{{y\textnormal{r}}} \right), \label{eq:psidotdot_in_dyn_model} \\
& \dot{v_x} = \frac{1}{m} F_x ,   \label{eq:vxdot_in_dyn_model} \\
& \dot{v_y} = \frac{1}{m} \left( F_{{y\textnormal{f}}} + F_{{y\textnormal{r}}} \right) - v_x \dot{\psi}, \label{eq:vydot_in_dyn_model}
\end{align}
\end{subequations}
where $s$ denotes the curvilinear abscissa i.e., the progression of the vehicle along the centerline of the lane with curvature $\kappa_c$ at $s$. Variable $d$ represents the normal distance from the centerline at $s$ to the center of mass of the vehicle. The variable $\Delta \psi$ denotes the vehicle orientation relative to the centerline tangent at $s$, and $\dot{\psi}$, $v_x$ and $v_y$ denote yaw rate, longitudinal and lateral velocities respectively. The inputs of the model are $F_{y\textnormal{f}}$, the lateral force on the front tire and $F_{x}$, the combined longitudinal force on the front and rear tires. The values $m$, $I_z$, $l_\textnormal{f}$ and $l_\textnormal{r}$ are physical vehicle parameters. For the purposes of this paper we assume that effects of longitudinal load transfer, bank angle and grade angle of the road are small. 
We compactly write (1) as $\dot{x} = f_c(x,u)$, where $x = [s \ d \ \Delta \psi \ \dot{\psi} \ v_x \ v_y]^\top$ and $u = [F_{y\textnormal{f}} \ F_x]^\top$. We then discretize (1) using forward Euler discretization, $x_{t+1} = x_t +  T_sf_c(x_t,u_t)$ with sampling time $T_s$, to get $x_{t+1} = f(x_t,u_t)$. 

\subsubsection*{Ideal Optimal Control Problem}
For trajectory planning and control synthesis, we wish to solve the following Constrained Finite Time Optimal Control (CFTOC) problem in a receding horizon \cite[Chapter 12]{borrelli2017predictive} fashion at any time instance $t$, for all $t \geq 0$:
\begin{equation}
\begin{array}{ll}
\underset{u_{0|t}, \cdots, u_{N-1|t}}{\mbox{min}} & p(x_{N|t})  + \displaystyle\sum_{k = 0}^{N-1}  q(x_{k|t},u_{k|t})   \\\
~~~~~~\mbox{s.t.} 	& x_{k+1|t} = f \left(x_{k|t},u_{k|t} \right), \\
			    & x_{k|t} \in \mathcal{X}_t, \ u_{k|t} \in \mathcal{U}_t(\mu_t), \\
			    & \forall~ k = 0,\ldots,N - 1,\\
			    & x_{0|t} = x_t,~x_{N|t} \in \mathcal{X}_t,  
\end{array}
\label{eq:cftoc}
\end{equation}
where $[x_{0|t},\cdots, x_{N|t}]^\top$  at time $t$, denote the predicted states along a prediction horizon of length $N$, when the predicted input sequence $[u_{0|t},\cdots,u_{N-1|t}]^\top$ is applied through vehicle model $f(\cdot,\cdot)$. The inputs $u_{k|t}$ for all $k \in \{0,\dots,N-1\}$ are bounded by the sets $\mathcal{U}_t(\mu_t) \subseteq \mathbb{R}^m$ to account for local dynamic limitations of the vehicle. $\mu_t$ is the identified tire-road friction coefficient. The sets $\mathcal{X}_t \subseteq \mathbb{R}^n$ represent collision free states of the vehicle with respect to drivable area, static and dynamic obstacles. Functions $q(\cdot, \cdot)$ and $p(\cdot)$ denote the positive definite running cost and terminal cost functions respectively. After solving \eqref{eq:cftoc} at each time $t$, the first optimal input $u_t = u^\star_{0|t}$ is to be applied in closed loop to \eqref{eq:dyn_model} and then \eqref{eq:cftoc} is to be solved at next time $t+1$, as per the receding horizon strategy. Notice that the problem \eqref{eq:cftoc} is formulated with time varying constraints $\mathcal{U}_t(\mu_t)$ and $\mathcal{X}_t$. Time variation in $\mathcal{X}_t$ is required to represent predicted movement of dynamic obstacles. Time variation in $\mathcal{U}_t(\mu_t)$ is introduced to account for variations in the physical capabilities of the vehicle due to \textcolor{black}{local traction variations.}

\subsubsection*{Adaptive Constraints}
\begin{figure}[ht]
	\centering
    \includegraphics[width=0.8\columnwidth]{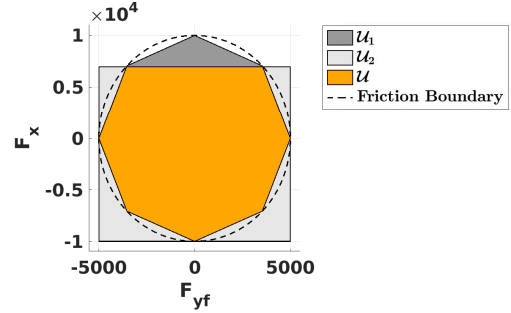}
    \caption{Adaptive input constraint polytope capturing local limitations of tire forces. The size of $\mathcal{U}(\mu)$ varies with the identified parameter $\mu$}
    \label{fig:inputconstraint}
\end{figure}

The maximum horizontal force that can be exerted between a tire and the road at time $t$ is determined by the normal force, $F_z$, and the tire-road friction coefficient $\mu_t$. The boundary of combined lateral and longitudinal forces on a single tire is referred to as a friction circle \cite{rajamani2011vehicle}. For our dynamic bicycle model we have that 
\begin{equation}
    F_{{y\textnormal{f}}}^2 + F_{{x\textnormal{f}}}^2  \leq (\mu F_{z{\textnormal{f}}})^2, \quad F_{{y\textnormal{r}}}^2 + F_{{x\textnormal{r}}}^2 \leq (\mu F_{z{\textnormal{r}}})^2.
    \label{eq:force_constraint}
\end{equation}
Considering the control inputs of \eqref{eq:dyn_model} and assuming that effects of longitudinal load transfer are small, the friction circle constraint is satisfied if the pair of control inputs $F_{{y\textnormal{f}}}$ and $F_x$ are inside an ellipse with half-axles $\mu F_{z{\textnormal{f}}}$ and $\mu F_z$. For computational tractability of \eqref{eq:cftoc}, we represent the input constraints as a set of affine constraints. For that reason, we determine a polytope $\mathcal{U}_1(\mu)$ that under-approximates the ellipse. Lower and upper bounds in $F_{{y\textnormal{f}}}$ and $F_x$ due to limits in steering angle and motor torque are represented as a second polytope $\mathcal{U}_{2}$. The final input constraint polytope is computed as the intersection $\mathcal{U}(\mu) = \mathcal{U}_{1}(\mu) \cap \mathcal{U}_{2} = \{u: H^\mu u \leq h^\mu\}$, illustrated in Fig.~\ref{fig:inputconstraint}.
In the next section, we use these parametric ($\mu$ dependent) adapted input constraints to formulate a trajectory planning and optimization algorithm, which attempts to solve \eqref{eq:cftoc} in real-time. 

\section{Sampling Augmented Adaptive RTI}
\label{sec:algo_main}
There are two practical problems with the ideal optimal control synthesis formulation in \eqref{eq:cftoc}, namely: 
\begin{enumerate}[(i)]
    \item A direct solution to \eqref{eq:cftoc} using a nonlinear solver would be prone to getting stuck in local minima, when the problem contains discrete decision making (e.g., to go left or right of an obstacle \cite{ziegler2014trajectory}). This can render the approach computationally intractable. 
    
    \item Even if \eqref{eq:cftoc} is tractably reformulated, due to the adaptive nature of the constraints, potential issues of feasibility in solving the reformulation of (2) might arise \cite{manrique2014mpc}, as a result of discrepancy between constraints $\mathcal{U}_t(\mu_t)$, $\mathcal{X}_t$ and dynamics $f(x_t,u_t)$.
\end{enumerate}
To address these problems, we propose the Sampling Augmented Adaptive RTI (SAA-RTI) algorithm, which decomposes the method of solving \eqref{eq:cftoc} into two distinct steps: \textit{feasible trajectory planning} and \textit{trajectory optimization}. The approach augments the existing RTI-SQP \cite{gros2016linear} strategy with state space sampling \cite{howard2008state}. The horizon and sampling time for both trajectory planning and optimization steps are chosen as $N$ and $T_s$ respectively, as in \eqref{eq:cftoc}.


For the \textit{feasible trajectory planning} step, we modify the state space sampling method in \cite{howard2008state} to handle the adaptive nature of actuation constraints (defined in Section~\ref{sec:problemformulation},~Fig.~\ref{fig:inputconstraint}), while satisfying vehicle dynamics \eqref{eq:dyn_model}. This incorporation of real time constraint adaptation, repeatedly generates a large set of feasible sampled trajectories at each time step. Unlike conventional RTI-SQP, the presence of this set
provides a systematic method for warmstarting the subsequent \textit{trajectory optimization}, remedying problem $(\text{i})$ defined above.

In the \emph{trajectory optimization} step, we define state constraints $\mathcal{X}_t$ w.r.t. deviations from feasible planned trajectories. As both trajectory planner and optimizer use the same vehicle model $f(x_t,u_t)$ and adapted constraints $\mathcal{U}_t(\mu_t)$, one planned trajectory is \emph{guaranteed} to be a feasible solution for the \textit{trajectory optimization} problem. Hence, $(\text{ii})$ is resolved. In the following two sub-sections we elaborate the details of these two  aforementioned steps  of the algorithm. 

\subsection{State Space Sampling and Trajectory Planning}
\label{sec:trajectoryselection}
To obtain a feasible and near-optimal solution to \eqref{eq:cftoc}, we first utilize state space sampling \cite{howard2008state}. The purpose of this step is to provide 
\textcolor{black}{additional, feasible} warmstarting options for the subsequent trajectory optimization problem, while being cognizant of time variations in operating constraints.  
At any time $t$, we first compute the initial state of the vehicle in the road aligned frame $[s_{0|t}, d_{0|t}, \Delta \psi_{0|t}, \dot{\psi}_{0|t}, (v_{x})_{0|t}, (v_{y})_{0|t}] \mapsto [s_{0|t}, d_{0|t}, \dot{s}_{0|t}, \dot{d}_{0|t}, \ddot{s}_{0|t}, \ddot{d}_{0|t} ]$. We then define a set of terminal states $[s_{N|t}, d_{N|t}, \dot{s}_{N|t}, \dot{d}_{N|t}, \ddot{s}_{N|t}, \ddot{d}_{N|t}]$, where $d_{N|t}$ denotes terminal lateral deviations from the lane centerline at $s_{N|t}$. Then, for each terminal state of the sampling based planner, a trajectory from the initial to the terminal state is determined by a piecewise affine function in $s$ and a quintic polynomial in $d$ by solving a set of two point boundary value problems. The coefficients of such a quintic polynomial $d_{k|t} = a_0 + a_1(\tilde{k}) + a_2(\tilde{k})^2 + a_3(\tilde{k})^3 + a_4(\tilde{k})^4 + a_5(\tilde{k})^5$ over the planning horizon of length $N$, i.e, $k \in \{0,\dots, N-1\}$ with $\tilde{k} = kT_s$ ($T_s$ is the sampling time defined in Section~\ref{sec:problemformulation}), can be efficiently computed by solving
\begin{equation*}
\begin{bmatrix}
1 & 0 & 0 & 0 & 0 & 0 \\
0 & 1 & 0 & 0 & 0 & 0 \\
0 & 0 & 1 & 0 & 0 & 0 \\
1 & \tilde{k} & \tilde{k}^2 & \tilde{k}^3 & \tilde{k}^4 & \tilde{k}^5 \\
0 & 1 & 2\tilde{k} & 3\tilde{k}^2 & 4\tilde{k}^3 & 5\tilde{k}^4 \\
0 & 0 & 2 & 6\tilde{k} & 12\tilde{k}^2 & 20\tilde{k}^3 
\end{bmatrix}
\begin{bmatrix}
a_0 \\
a_1 \\
a_2 \\
a_3 \\
a_4 \\
a_5
\end{bmatrix}
=
\begin{bmatrix}
d_{0|t} 	\\
\dot{d}_{0|t} 	\\
\ddot{d}_{0|t} 	\\
d_{N|t} 	\\
\dot{d}_{N|t} 	\\
\ddot{d}_{N|t} 	\\
\end{bmatrix}.
\end{equation*}
Computation of coefficients for piecewise affine $s_{k|t},~k\in \{0,\dots,N\}$ is trivial, and is omitted due to limited space. 

Each trajectory from the planner is then transformed to a state trajectory of the vehicle, satisfying dynamics \eqref{eq:dyn_model} smoothly. This transformation $\hat{\gamma} = [s, d, \dot{s}, \dot{d}, \ddot{s}, \ddot{d} ] \mapsto \hat{x} = [s, d, \Delta \psi, \dot{\psi}, v_{x}, v_{y}]$ is computed in closed form. Calculations are available in \cite{werling2012_IJRRl}.
Following the above transformation $\hat{\gamma} \mapsto \hat{x}$, we loop through the planned trajectory set at any time $t$ over a finite prediction horizon of length $N$, checking feasibility of dynamic constraints (see Fig.~\ref{fig:inputconstraint}) and collision avoidance $\mathcal{X}_t$ . \textcolor{black}{The adaptive force constraints are checked by computing the equivalent tire forces of each trajectory and checking them with respect to constraints $\mathcal{U}_t(\mu_t)$.} For trajectories passing this check, we evaluate a cost metric (essentially the to cost of \eqref{eq:cftoc}. See Remark~\ref{rem:J} for details)
\begin{equation}
    \label{eq:cost}
    J(\hat{X}_{t}) = \hat{x}_{N|t}^\top Q_f \hat{x}_{N|t}  + \displaystyle\sum_{k=0}^{N-1} \hat{x}_{k|t}^\top Q \hat{x}_{k|t} + \hat{u}_{k|t}^\top R \hat{u}_{k|t},
\end{equation}
where $\hat{X}_t = \{(\hat{x}_{k|t},\hat{u}_{k|t}),~k \in \{0,\dots,N\}\}$ denotes a planned trajectory (suboptimal) rolled out by the state-space sampling planner, \textcolor{black}{which satisfies the vehicle dynamics $\hat{x}_{k+1|t} = f(\hat{x}_{k|t},\hat{u}_{k|t}),~k \in \{0,\dots,N\}$}. The matrices $Q, Q_f, R \succ 0$ are \textcolor{black}{tuning matrices}, selected such that the cost reflects the overall objective. 
The lowest cost sampled trajectory at $t=0$ i.e. $\hat{X}^\star _0 = \{(\hat{x}^\star_{k|0},\hat{u}^\star _{k|0}),~k \in \{0,\dots,N\}\}$ (where $\hat{X}^\star_0 = \arg \min \limits_{{X}} J(\hat{X}_0) = (\hat{x}^\star_{k|0},\hat{u}^\star _{k|0}),~k \in \{0,\dots, N\}$) is selected for the subsequent trajectory optimization at time $t=0$ to obtain the optimal $X^\star_0 = \{(x^\star_{k|0},u^\star_{k|0}),~ k \in \{0,\dots,N\}\}$. From $t=1$ onward, $\hat{X}^\star_{t} = \arg \min \limits_{{X}} (J(\hat{X}_{t}), J(X^\star_{t-1}))$. That is, the forward shifted optimal trajectory $X^\star_{t-1} = \{ (x^{\star }_{k|t-1},u^{\star }_{k|t-1}),~k \in \{1,\dots,N\}\}$ from the previous iteration of the trajectory optimization is included in the selection on equal terms with the sampled trajectories $\hat{X}_t$ for the current time step. 
\begin{remark}
In case $X^\star_{t-1}$ is selected as $\hat{X}^\star_t$, the algorithm behaves as standard RTI-SQP \cite{gros2016linear}.
\end{remark}

Although dynamically feasible and optimal within the sampled set, $\hat{X}^\star_t$ will be suboptimal due to the structure imposed by the polynomials defining the trajectory. We will later illustrate this in Section \ref{seq:resultsanddiscussion}. Hence, we employ trajectory optimization using RTI-SQP to obtain an optimal trajectory $X^\star_t$ from the initialized suboptimal trajectory $\hat{X}^\star_t$ for all $t\geq 0$, rather than simply tracking suboptimal $\hat{X}^\star_t$.


\subsection{Trajectory Optimization}
\label{sec:trajectoryoptimization}
As stated in problem $(\text{i})$ in the beginning of this section, the ideal adaptive CFTOC problem \eqref{eq:cftoc} is non-convex, so \textcolor{black}{solvers are} prone to getting stuck in local minima. However, solving \eqref{eq:cftoc} locally around the feasible but suboptimal trajectory $\hat{X}^\star_t$ can be done efficiently using a convex Quadratic Program (QP) approximation. We obtain the QP approximation of \eqref{eq:cftoc} through the linear time varying model predictive control paradigm \cite{gros2016linear}. At any given time $t$, the model and constraints in \eqref{eq:cftoc} are linearized around $\hat{X}^\star_t$. Then, for one iteration of the algorithm, the following reformulated optimization problem is solved once at each time step $t$, instead of solving \eqref{eq:cftoc}: 

\begin{equation}\label{eq:mpc_problem}
\begin{array}{ll}
\! \! \! \underset{\Delta u_{0|t},\dots, \Delta u_{N-1|t}}{\mbox{min}} & \! \! \! \! J(x_{k|t},u_{k|t}) + \sigma_t^\top \beta \sigma_t \\\
~~~~~\mbox{s.t.} 	& x_{k+1|t} \! = \! A_{k|t}(\Delta x_{k|t}) \! + \! B_{k|t} (\Delta u_{k|t}) \! + \! \hat{x}^\star_{k+1|t}, \\
			    & H^\mu_t u_{k|t} \leq h^\mu_t, \\
			    & \forall ~ k = 0,\ldots,N - 1,~\textnormal{and, }  \vspace{2mm} \\ 
			    & s^{\textnormal{min}}_{k|t} - \sigma^s_t \leq s_{k|t} \leq s^{\textnormal{max}}_{k|t} + \sigma^s_t, \\
			    & d^{\textnormal{min}}_{k|t} - \sigma^d_t \leq d_{k|t} \leq d^{\textnormal{max}}_{k|t} + \sigma^d_t, \\
			    & (v^{\textnormal{min}}_{x})_{k|t} - \sigma^{v_x}_t \leq (v_{x})_{k|t} \leq (v^{\textnormal{max}}_{x})_{k|t} + \sigma^{v_x}_t, \\
			    & \forall ~ k = 0,\ldots,N,  \\
			    & x_{0|t} = x_t, \\
			    & \sigma^s_t \geq 0, \ \sigma^d_t \geq 0, \ \sigma^{v_x}_t \geq 0, 
\end{array}
\end{equation}
where $[x_{1|t},\cdots, x_{N|t}]$ are predicted states obtained in open loop at time $t$, after applying the predicted input sequence $[u_{0|t},\cdots, u_{N-1|t}]$ to the linearized system, and $[\Delta x_{k|t}, \Delta u_{k|t}] = [x_{k|t} - \hat{x}^\star_{k|t}, u_{k|t}-\hat{u}^\star_{k|t}]$ for all $k \in \{0,\dots,N-1\}$. The linearized system model matrices are given as 
$ A_{k|t} = \frac{\partial f}{\partial x}  \bigg|_{(\hat{x}^\star_{k|t}, \hat{u}^\star_{k|t})}, \quad 
    B_{k|t} = \frac{\partial f}{\partial u}  \bigg|_{(\hat{x}^\star_{k|t}, \hat{u}^\star_{k|t})}$,
\vspace{0.1cm}
for all $k \in \{0,\dots,N-1\}$ and state constraints $\hat{\mathcal{X}}_{k|t} = \{ s^{\textnormal{min}}_{k|t} - \sigma^s_t \leq s_{k|t} \leq s^{\textnormal{max}}_{k|t} + \sigma^s_t, ~~ d^{\textnormal{min}}_{k|t} - \sigma^d_t \leq d_{k|t} \leq d^{\textnormal{max}}_{k|t} + \sigma^d_t, ~~ (v^{\textnormal{min}}_x)_{k|t} - \sigma^{v_x}_t \leq (v_{x})_{k|t} \leq (v^{\textnormal{max}}_x)_{k|t} + \sigma^{v_x}_t \} $ for all $k \in \{0,\dots,N\}$ in \eqref{eq:mpc_problem} are selected, such that the deviation from $\hat{X}^\star_t$ is bounded. The constraints are softened with slack variables $\sigma_t$ to maintain feasibility of \eqref{eq:mpc_problem} and any constraint violation is heavily penalized by $\beta \gg 0$.  
The same cost function $J(\cdot)$ as in the trajectory selection (shown in \eqref{eq:cost}) is employed, with the addition of the term $\sigma_t^\top \beta \sigma_t$, with $\sigma_t = [\sigma^s_t, \sigma^d_t, \sigma^{v_x}_t]^\top$, to account for slack variables in the soft state constraints. 
After solving \eqref{eq:mpc_problem}, we apply the first input $u_t = u^\star_{0|t}$. 

\begin{remark}\label{rem:J}
WLOG in \eqref{eq:cftoc}, we choose $p(x_{N|t})  + \sum \limits_{k = 0}^{N-1} q(x_{k|t},u_{k|t}) = J(x_{k|t},u_{k|t})$ for \eqref{eq:cost} and \eqref{eq:mpc_problem}. 
\end{remark}

We highlight that the novelty in \eqref{eq:mpc_problem} is two fold:
\begin{enumerate}
    \item Inclusion of the adaptive constraint polytope $\mathcal{U}(\mu_t) = \{u: H^\mu_t u \leq h^\mu_t\}$, where $\mathcal{U}(\mu_t)$ is recomputed at every time $t$ from the identified parameter $\mu_t$ as per the method described in Section~\ref{sec:problemformulation}, Fig.~\ref{fig:inputconstraint}. This enables the resulting optimal trajectory $X^\star_t$ to fully utilize the available tire force, given the current driving conditions. This improves the vehicle's capacity to avoid obstacles.
    
    \item Inclusion of sampled trajectories $\hat{X}_t$ to warmstart the optimization problem \eqref{eq:mpc_problem} at each time $t$. This alleviates the issue of local minima and potential infeasibility of \eqref{eq:mpc_problem}. Since the same vehicle model \eqref{eq:dyn_model} and adaptive constraints (given by $\mathcal{U}_t(\mu_t)$) are applied in both trajectory selection (Section~\ref{sec:trajectoryselection}) and the trajectory optimization (Section~\ref{sec:trajectoryoptimization}) steps, one feasible solution to \eqref{eq:mpc_problem} is guaranteed to exist at any time $t$, namely $\hat{X}^\star_t$.
\end{enumerate}

We summarize the proposed Sampling Augmented Adaptive RTI (SAA-RTI) algorithm in Algorithm~1. Let $M$ represent the map features, e.g., lane boundaries and static obstacles, and $O$ denote dynamic obstacles. At any time $t$, we assume an existing tire-road friction estimate  \cite{rajamani2012algorithms,gustafsson1997slip} $\mu_t$, and $\mathcal{T}_t$ (defined as $\cup_{i=1}^{\infty} \hat{X}^i_t)$ denotes the set of sampled trajectories. 

\begin{algorithm}[H] \label{alg:saa-rti}
    \caption{The SAA-RTI Algorithm}
    \begin{algorithmic}[1]
    \renewcommand{\algorithmicrequire}{\textbf{Input:}}
    \renewcommand{\algorithmicensure}{\textbf{Output:}}
    \REQUIRE $x_t$, $X^\star_{t-1}$, $M$, $O$
    \ENSURE $X^\star_t$               
    \STATE $\mu_t$ $\leftarrow$ \text{identifyFrictionCoefficient($x_t$)}
    \STATE $\mathcal{U}_t(\mu_t)$ $\leftarrow$ \text{computeAdaptiveConstraints}($\mu_t$)
    \STATE $\mathcal{T}_t$ $\leftarrow$ \text{sampleStateTrajectories}($x_t, M$)
    \FOR {each trajectory $\hat{X}_t^i$ in $[\mathcal{T}_t, X^\star_{t-1}]$} 
    \IF {(\text{chkConstr}($\hat{X}_t^i, \mathcal{U}_t(\mu_t) $) $\land$ \text{chkColl}($\hat{X}_t^i, O$))}
    \STATE $J(\hat{X}_t)$ $\leftarrow$ \text{evaluateCost}($\hat{X}_t^i$)
    \ENDIF
    \ENDFOR
    \STATE $\hat{X}^\star_t$ $\leftarrow$ \text{selectLowestCost}($\arg J(\hat{X}^\star_t)$)
    \STATE $A_{k|t}, B_{k|t}$ $\leftarrow$ \text{linearizeDynamicModel}($\hat{X}^\star_t$)
    \STATE $\hat{\mathcal{X}}_{k|t}$ $\leftarrow$ \text{computeStateConstraints}($\hat{X}^\star_t$, $O$, $M$) 
    \STATE $X^\star_{t} =  (x^{\star }_{k|t},u^{\star }_{k|t})$ $\leftarrow$ \text{opti}($\hat{X}^\star_t, \mathcal{U}_t(\mu_t), \hat{\mathcal{X}}_{k|t}, A_{k|t}, B_{k|t}$)
    \RETURN $X^\star_t$           
    \end{algorithmic}
\end{algorithm}

\section{Results and Discussion}
\label{seq:resultsanddiscussion}

In this section via thorough numerical analysis we demonstrate two aspects of Algorithm~1: 

\begin{enumerate}[(a)]
    \item First, we compare adaptive (real time adaptation of constraints $\mathcal{U}_t(\mu_t)$ defined in Section~\ref{sec:problemformulation}) and non-adaptive trajectory optimization by evaluating the realized closed loop cost ${J}_{\textnormal{cl}}(x_0) = \sum \limits_{t=0}^{\infty} \{ J(x_t,u^\star_{0|t}) + \sigma^{\star \top}_t \beta \sigma^\star_t \} $. \textcolor{black}{We generalize the results using Monte Carlo simulations by computing its empirical mean $\bar{J}_\textnormal{cl}$}, and empirical probability \textcolor{black}{of colliding or veering off the road}, denoted as $\mathbb{P}_{\textnormal{acc}}$.
    
    \item Second, we highlight the advantage of SAA-RTI for adaptive trajectory planning and optimization compared to state-of-the-art methods in terms of optimality and the ability to avoid local minima.
\end{enumerate}
Simulations are conducted in closed loop with a nonlinear bicycle model with model parameters stated in Table \ref{tab:static_params}. The resulting quadratic programs are solved with the Gurobi 
solver package in MATLAB. 

\begin{table}[t] 
    \centering
    \begin{tabular}{|c|c|}
        \hline
        \bfseries Parameter & \bfseries Value\\
        \hline\hline
        $m$ & 1500 kg \\
        $I_z$ & 2250 $\text{kgm}^2$ \\
        $l_{\textnormal{f}}$ & 1.04 m \\
        $l_{\textnormal{r}}$ & 1.42 m \\
        $C_{\alpha_\textnormal{f}}$ & 160 kN/rad \\
        $C_{\alpha_\textnormal{r}}$ & 180 kN/rad \\
        \hline
    \end{tabular}
    \vspace{0.2cm}
    \caption{Static parameters of dynamic vehicle model used in simulations}
    \label{tab:static_params}
    \vspace{-10mm}
\end{table}

\subsection{Adaptive vs. Non-Adaptive Trajectory Optimization}
\label{sec:adaptive_vs_nonadaptive}
Given the unavoidable local variations in the actual tire-road friction $\mu_{\textnormal{act}}$, a non-adaptive motion planning strategy will use (explicitly or implicitly) an assumed friction coefficient $\mu_{\textnormal{asm}}$, that at times differs significantly from $\mu_{\textnormal{act}}$. 
In \textcolor{black}{our first} evaluation scenario, the vehicle is driving on a curved section of road at a velocity of 15 m/s. An obstacle, which we assume is stationary, appears suddenly 15 meters ahead of the vehicle. Once the critical situation is detected, the goal of the vehicle is to reduce its speed to zero as soon as possible and come to a halt, while avoiding collision with high probability. 
The road conditions are divided in two cases: wet road, with  $\mu_{\textnormal{act}} = 0.55$ and dry road, with $\mu_{\textnormal{act}} = 0.95$. The non-adaptive trajectory optimization assumes a static friction estimate $\mu_{\textnormal{asm}} = 0.8$ throughout, while our proposed SAA-RTI re-estimates this value and accordingly adapts input constraints $\mathcal{U}_t(\mu_t)$ (see Fig.~\ref{fig:inputconstraint}) in \eqref{eq:mpc_problem}. \textcolor{black}{To mimic the convergence time of a friction estimation algorithm, we introduce a time delay of 100ms before the correct value of $\mu$ is applied in SAA-RTI.}

\subsubsection*{Adapting to Lower Traction}
\label{sec:adapting_to_lower_traction}
\begin{figure}[t]
\centering
    \subfloat[Wet road, not adapting, $J_{\textnormal{cl}} = 7.00$ \label{fig:wet_road_not_adapting}]{%
        \includegraphics[width=\columnwidth]{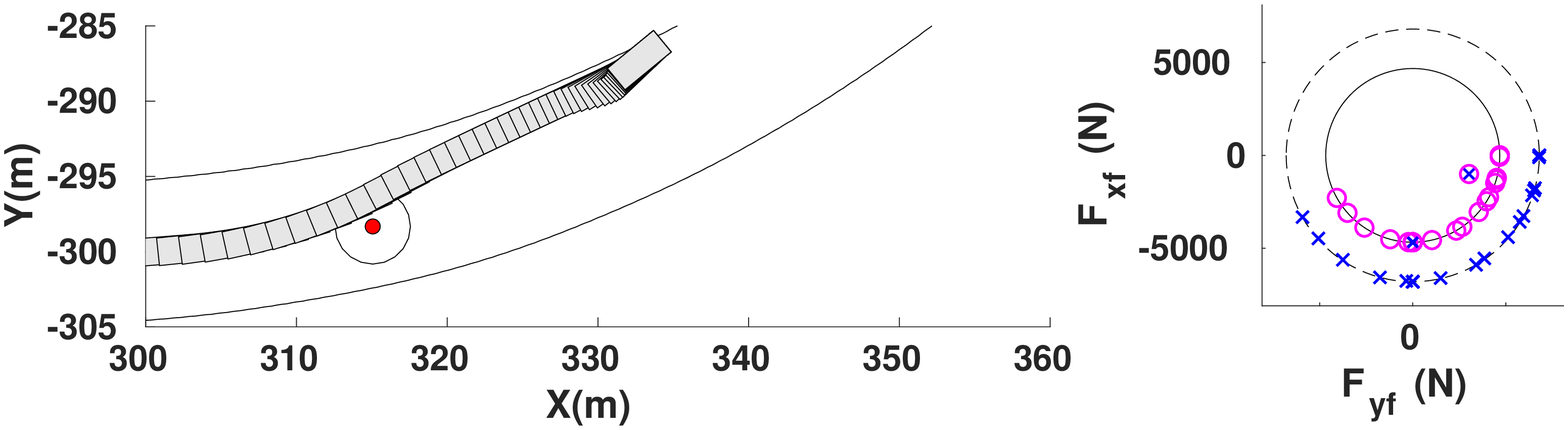}
    }\\
    \vspace{4pt}
    \subfloat[Wet road, adapting, $J_{\textnormal{cl}} = 6.71$ \label{fig:wet_road_adapting}]{%
        \includegraphics[width=\columnwidth]{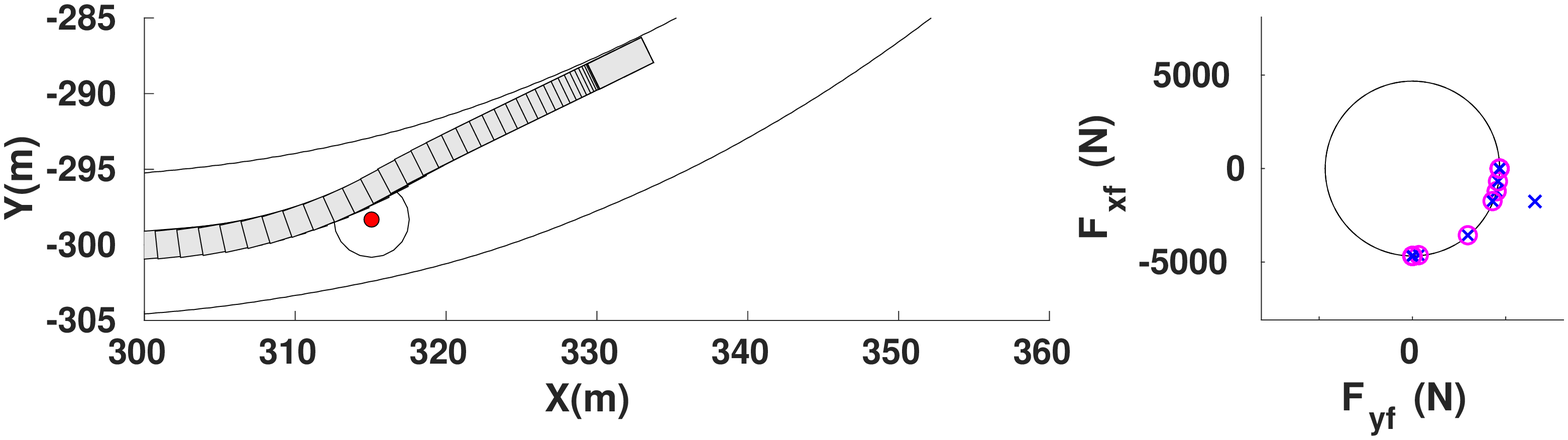}
    }\\
    \vspace{4pt}
    \subfloat[Dry road, not adapting , $J_{\textnormal{cl}} = 4.80$ \label{fig:dry_road_not_adapting}]{%
        \includegraphics[width=\columnwidth]{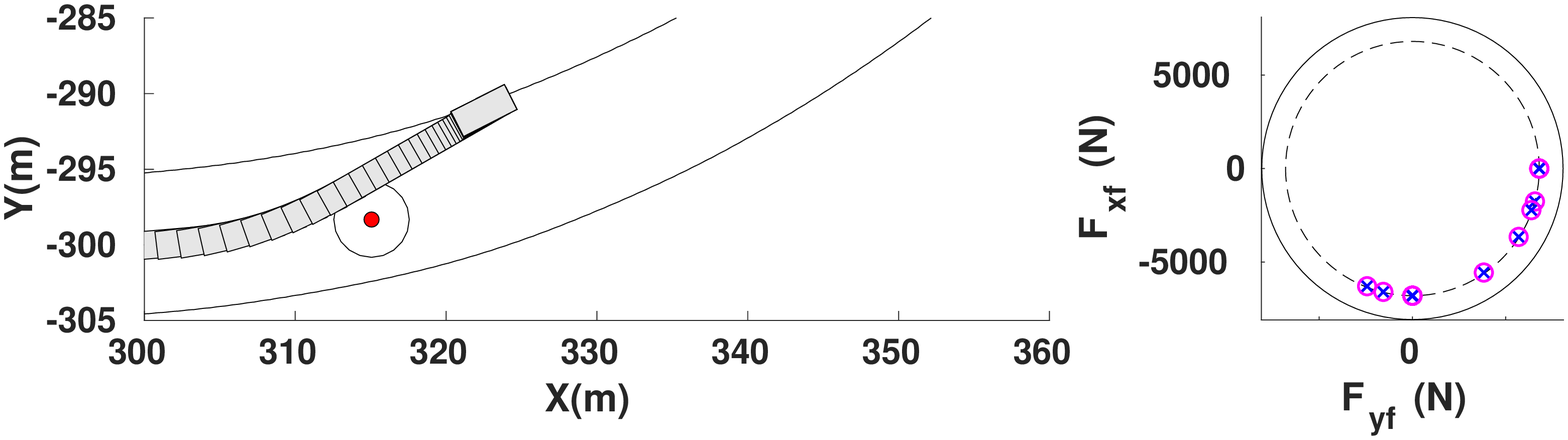}
    }\\
    \vspace{4pt}
    \subfloat[Dry road, adapting, $J_{\textnormal{cl}} = 4.08$ \label{fig:dry_road_adapting}]{%
        \includegraphics[width=\columnwidth]{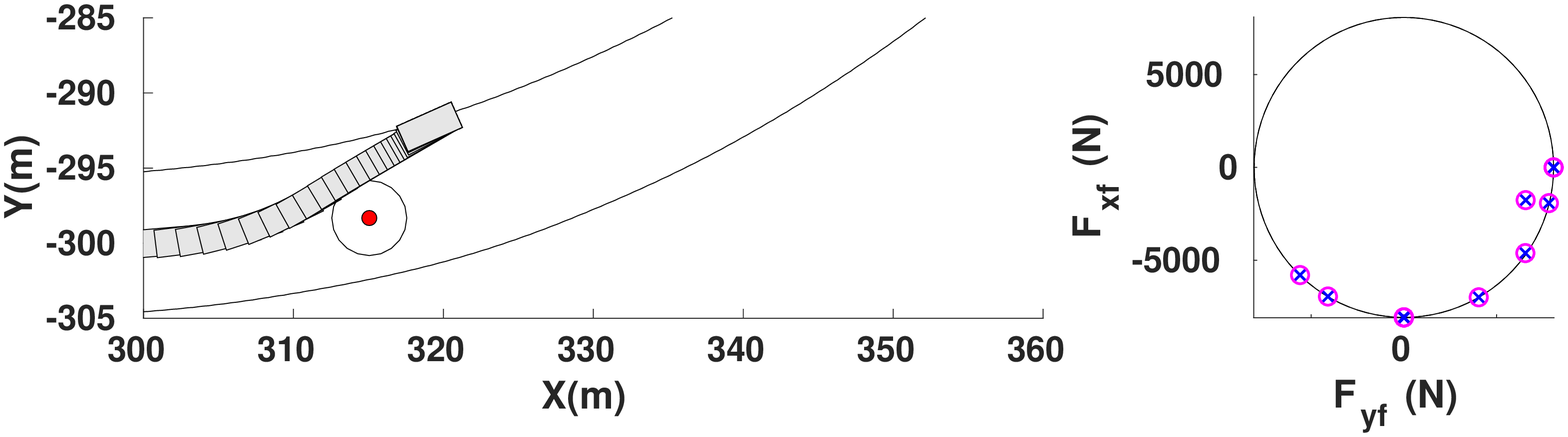}
    }
    \vspace{2pt}
    \caption{Closed loop trajectories for comparison between adaptive and     non-adaptive trajectory planning and control. The vehicle is depicted in gray, the suddenly appearing obstacle in red. In the force plot to the right, blue crosses denote the commanded tire forces and magenta circles denote actual tire forces. The solid and dashed black lines represent the actual and assumed friction circles respectively.}
\vspace{-15pt}
\end{figure}

First we compare non-adaptive, Fig.~\ref{fig:wet_road_not_adapting}, vs. adaptive, Fig.~\ref{fig:wet_road_adapting}, trajectory optimization in the case where the actual traction is \textit{below} the default assumption, $\mu_{\textnormal{act}} < \mu_{\textnormal{asm}}$. The realized closed loop cost $J_{\textnormal{cl}}$ starting from the same initial state is comparable in the two cases ($4\%$ difference), but in the non-adaptive case, the vehicle develops notably more side slip during the maneuver. The underlying cause for this is that the vehicle is unable to realize the planned motions due to saturated tire forces, as constraints are not adapted to match actual road conditions. Fig.~\ref{fig:wet_road_not_adapting} (right) shows the discrepancy between commanded tire forces (blue) and real tire forces (magenta) and the resulting sliding motion of the vehicle (left). In Fig.~\ref{fig:wet_road_adapting}, we see a notably lower discrepancy due to real time re-estimation of $\mu_{\textnormal{asm}}$ and subsequent constraint adaptation in SAA-RTI. The associated trajectory indicates that in this case, adapting gives reduced side-slip and enhanced stability during the evasive maneuver.

\subsubsection*{Adapting to Higher Traction} 
\label{sec:adapting_to_higher_traction}
Here we compare non-adaptive Fig.~\ref{fig:dry_road_not_adapting}, vs. adaptive, Fig.~\ref{fig:dry_road_adapting}, trajectory optimization in the case where the actual traction is \textit{above} the default assumption $\mu_{\textnormal{act}} > \mu_{\textnormal{asm}}$. 
We note that in this case, adapting decreases the stopping distance and the velocity at which the obstacle is passed, which is reflected by a 15\% decrease in $J_{\textnormal{cl}}$. The cause for this is evident from the front tire force plot in the right part of Fig.~\ref{fig:dry_road_not_adapting}. It reveals that the commanded tire forces (blue) are saturated by the friction circle associated with $\mu_{\textnormal{asm}}$ (dashed black), and therefore the vehicle is unable to fully utilize the available tire force without adaptation. In Fig.~\ref{fig:dry_road_adapting}, we see that constraint adaptation remedies the undesired saturation of commanded tire forces, resulting in a quicker, safer maneuver.   

\subsubsection*{Monte Carlo Analysis}

\textcolor{black}{We investigate the generality of the above indications by performing $1200$ Monte Carlo simulations of varying scenarios. For three different initial conditions (20m/s straight road, 15m/s curved road, 10m/s tight curved road), an obstacle appears at a random position in front of the vehicle. We compute performance metrics $\Bar{J}_\textnormal{cl}$ (i.e average closed loop cost of non-colliding runs) and $\mathbb{P}_{\textnormal{acc}}$, the empirical probability over all runs of colliding or veering off the road.}
\begin{table}[t]
\renewcommand{\arraystretch}{1.3}
\caption{\textcolor{black}{Results from Monte Carlo simulations of 1200 critical obstacle avoidance scenarios with varying initial conditions, obstacle positions, road conditions and control strategies. $\Bar{J}_\textnormal{cl}$ denotes the average closed loop cost over non-colliding trajectories and $\mathbb{P}_{\textnormal{acc}}$ the probability of colliding or exiting the road over all runs. For the non-adaptive case, $\mu_{\textnormal{asm}} = 0.8$.}}
\vspace{2mm}
\label{tab:mc_results}
\centering
\begin{tabular}{c|c|c|c}
\hline
Road Conditions & Strategy & $\Bar{J}_\textnormal{cl}$ & $\mathbb{P}_{\textnormal{acc}}$  \\ 
\hline
\hline
&  non-adaptive   & 5.33   &    42\%  \\  [-1ex]
\raisebox{1.5ex}{wet road: $\mu_{\textnormal{act}} = 0.55$} &  adaptive       & 5.37   &    38\%  \\  
\hline
&  non-adaptive   & 3.84   &    13\%  \\  [-1ex]
\raisebox{1.5ex}{dry road: $\mu_{\textnormal{act}} = 0.95$} &  adaptive       & 3.37   &    9\%  \\  
\hline
\end{tabular}
\vspace{-6mm}
\end{table}

\textcolor{black}{The results presented in Table~\ref{tab:mc_results} show that traction adaptive motion planning and control, improves the capacity to avoid accidents both in dry and wet road conditions, by enabling full utilization of available tire forces without loss of control authority.}

\subsection{Optimality and Feasibility of SAA-RTI}

\begin{figure}[t]
    \centering
     \includegraphics[width=1.0\columnwidth]{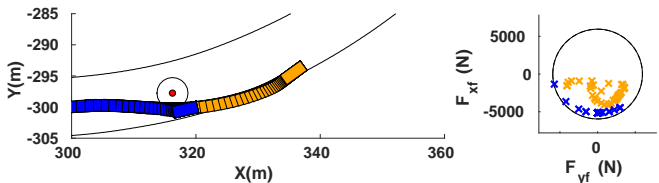} \\
    \caption{Comparison of closed loop trajectories between SAA-RTI (blue) and SSS-MPC (orange)}
    \label{fig:saspq_vs_sssmpc}
\end{figure}

In order to demonstrate \textcolor{black}{the quality}
of realized trajectories, we compare SAA-RTI with a standard modular approach with separated trajectory planning and tracking \cite{gonzalez2016review}, which uses state space sampling of Section~\ref{sec:trajectoryselection} for planning and MPC for tracking the planned trajectory (which is $\arg \min \limits_{X} J(\hat{X}_t)$). We refer to the method as State Space Sampling with MPC tracking (SSS-MPC). The key difference in the two approaches is that instead of tracking a suboptimal $\arg \min \limits_{X} J(\hat{X}_t)$ from the planner, the SAA-RTI optimizes the selected trajectory (see Section~\ref{sec:trajectoryselection} and Algorithm~1) to obtain an optimal trajectory (${X}^\star_t$,~see Section~\ref{sec:trajectoryoptimization} and Algorithm~1). As a result, in Fig.~\ref{fig:saspq_vs_sssmpc} we see that SAA-RTI stops in a shorter distance and passes the obstacle at a lower speed, which is reflected by a decrease in average closed loop cost $\Bar{J}_{\textnormal{cl}}$, by 42.2 \% over 100 runs with a randomly instantiated obstacle.

\begin{figure}[t]
    \centering
     \includegraphics[width=0.7\columnwidth]{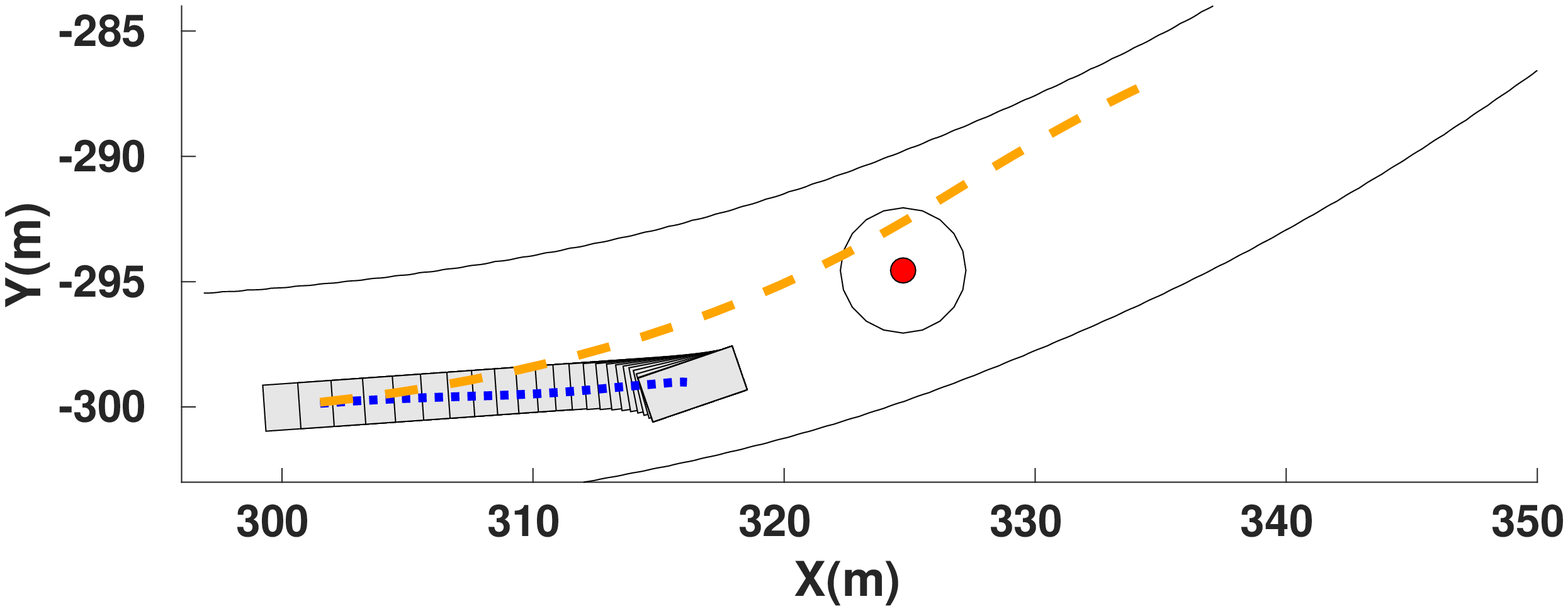} \\
    \caption{Example of how SAA-RTI avoids local minima. Orange: converged SQP solution initialized left of obstacle. Blue: converged SQP solution initialized right of obstacle. Gray: Closed loop trajectory of the vehicle controlled by SAA-RTI}
    \label{fig:local_minima}
    \vspace{-11pt}
\end{figure}

Moreover, a direct solution of \eqref{eq:cftoc} using RTI-SQP is sensitive to local minima. This phenomenon is highlighted in Fig.~\ref{fig:local_minima}. The fully converged SQP solution initialized to the left of the obstacle has a significantly higher cost, $7.94$, than that of the solution initialized to the right of the obstacle, which is $3.05$. Thus, the solution to the left of the obstacle constitutes a local minima of \eqref{eq:cftoc}. We observe from Fig.~\ref{fig:local_minima} that SAA-RTI makes the discrete decision to go right of the obstacle and has a closed loop cost $3.156$ close to the global optimum, avoiding the local minimum. 


\section{Conclusions}
\label{sec:conclusionsandfuturework}

In order to make full use of the physical capacity of an automated vehicle to avoid collisions in critical scenarios, we propose an integrated framework for trajectory planning and optimization that adapts to current traction limitations. By updating information on operating conditions
\textcolor{black}{in the integrated planning and optimization framework,}
we ensure safe constraint adaptation and feasible trajectory generation at the limits of handling. 
We demonstrate that traction adaptive trajectory planning improves the capacity to avoid accidents
\textcolor{black}{by fully utilizing the available tire forces, while maintaining control authority of the vehicle.}
%
Furthermore, \textcolor{black}{by augmenting Real Time Iteration-Sequential Quadratic Programming with} state space sampling, 
our proposed \textcolor{black}{optimization based planning-control} algorithm called SAA-RTI, delivers an improvement in terms of feasibility and optimality, demonstrated with thorough Monte Carlo simulations. 

\section*{ACKNOWLEDGMENT}
The authors gratefully acknowledge the AutoDrive project, H2020-ECSEL and Hyundai Center of Excellence at UC Berkeley, for financial support.

\balance 

\bibliographystyle{IEEEtran}
\bibliography{references}


\end{document}